\documentclass[runningheads]{llncs}
\usepackage{graphicx}
\usepackage{tikz}
\usepackage{comment}
\usepackage{amsmath,amssymb} % define this before the line numbering.
\usepackage{color}
\usepackage{hyperref}

\usepackage[singlelinecheck=false]{caption}
\usepackage{listings} % TAKE OUT AND USE VERBATIM

\begin{document}
\title{PirouNet: Creating Dance through Artist-Centric Deep Learning}
\titlerunning{PirouNet: Creating Dance through Artist-Centric Deep Learning}
%\author{Anonymous Submission}
\author{Mathilde Papillon\inst{1}\orcidID{0000-0003-1674-4218} \and
Mariel Pettee\inst{2}\orcidID{0000-0001-9208-3218} \and
Nina Miolane\inst{3}\orcidID{0000-0002-1200-9024}}
\authorrunning{M. Papillon et al.}
\institute{University of California Santa Barbara, Department of Physics, Santa Barbara, CA, USA \and
Lawrence Berkeley National Lab, Berkeley, CA, USA \and University of California Santa Barbara, Department of Electrical and Computer Engineering, Santa Barbara, CA, USA }
\maketitle              % typeset the header of the contribution
\begin{abstract}
Using Artificial Intelligence (AI) to create dance choreography  is still at an early stage. Methods that conditionally generate dance sequences remain limited in their ability to follow choreographer-specific creative direction, often relying on external prompts or supervised learning. In the same vein, fully annotated dance datasets are rare and labor intensive. To fill this gap and help leverage deep learning as a meaningful tool for choreographers, we propose ``PirouNet", a semi-supervised conditional recurrent variational autoencoder together with a dance labeling web application. PirouNet allows dance professionals to annotate data with their own subjective creative labels and subsequently generate new bouts of choreography based on their aesthetic criteria. Thanks to the proposed semi-supervised approach, PirouNet only requires a small portion of the dataset to be labeled, typically on the order of 1\%. We demonstrate PirouNet's capabilities as it generates original choreography based on the ``Laban Time Effort", an established dance notion describing a given intention for a movement's time dynamics. We extensively evaluate PirouNet's dance creations through a series of qualitative and quantitative metrics, validating its applicability as a tool for choreographers.

\end{abstract}
\section{Introduction}
Recent years have witnessed the development of deep learning methods that can generate original dance sequences. Yet, these methods have not been widely adopted by communities of dancers and choreographers. A possible reason for this lack of adoption is the fact that existing AI-powered dance generation tools cannot create dance based on artistic criteria, or ``user-inputs" that are meaningful to choreographers, such as elements of Laban Movement Analysis \cite{laban_book1,groff1995laban}. 

\paragraph{Laban Movement Analysis (LMA)} \cite{laban_book1,groff1995laban} stands as one of the most established Western methods for describing and understanding human motion \cite{laban_mvt_dance_concepts}, often used by dance professionals to transmit choreography, teach technique \cite{laban_for_ballet,childrensdance}, and rigorously assess performance \cite{lma_folk_2015}. LMA's corresponding notation, Labanotation (LN), enables detailed and objective recording of the quantitative and qualitative characteristics of movement \cite{labanotation}. As a result, LMA has become a cross-disciplinary staple of technologically-driven movement analysis and has been extensively assessed for reliability and effectiveness \cite{assess_laban,perceptual_cons_laban,seeing_laban}. LMA now provides a valuable tool for many movement-centric fields beyond dance, such as physical rehabilitation \cite{laban_book1}, theater \cite{drama_therapy}, sports \cite{hamburg_coaching}, and psychology \cite{somatic,neurology,behavior}.

\begin{figure}
	\centering
 	\includegraphics[width=0.9\linewidth]{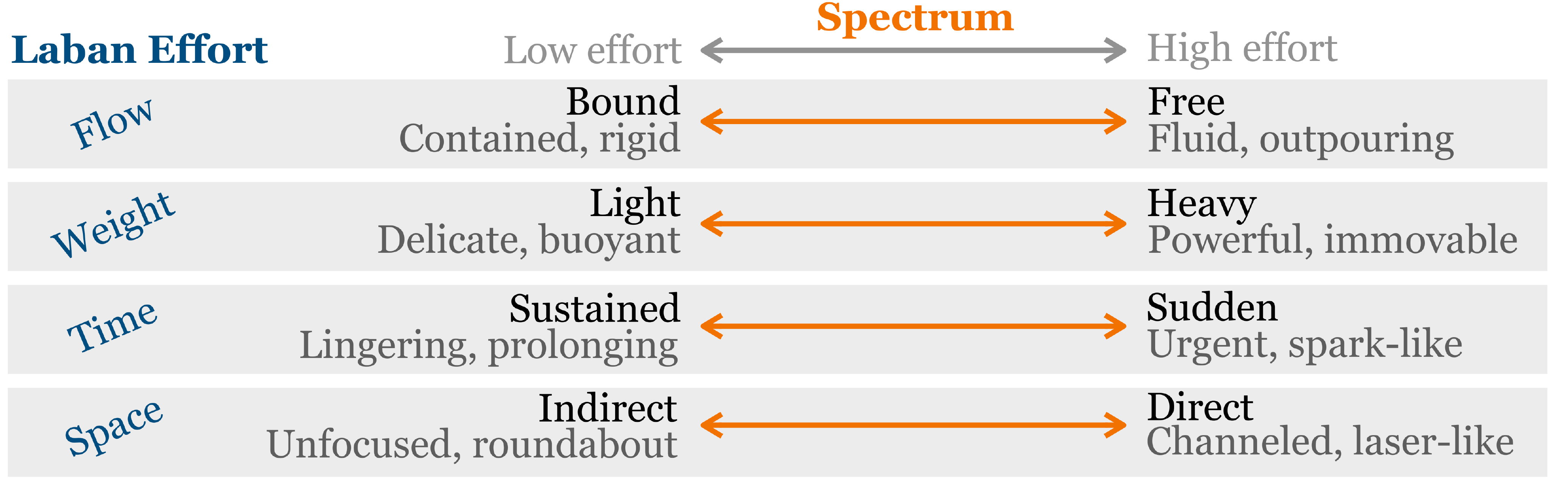}
	\caption{LMA's Effort axis consists of a four-dimensional space spanned by four orthogonal efforts called Laban Efforts. For every movement, there exists a 4-vector of total effort. For example, a movement that is described as bound, light, sustained, and direct has a total effort comprised of Low Flow Effort, Low Weight Effort, Low Time Effort, and High Space Effort. }
	\label{fig:efforts}
\end{figure}

\paragraph{Laban Axes and Laban Efforts.} LMA is divided into four axes which independently contribute to its portrayal of movement \cite{laban_book2_2006}. ``Body" outlines actions of body parts and relationships between those parts, ``Shape" addresses the body’s more general evolution of shape, ``Space" describes the geometrical and directional emphasis of movement, and ``Effort" describes the energy content and inner intention of a movement. In the context of this paper, we use the Effort axis as a leading example of aesthetic criterion, depicted in Fig. \ref{fig:efforts}, for AI-powered dance creation. We describe choreographic intention as the intended Effort of a movement. Effort is a multi-dimensional axis of LMA, described by four independent Laban Efforts. It truly sets LMA apart, as it describes an intention of movement, rather than the resulting execution \cite{def_efforts,robot_efforts}.

\paragraph{PirouNet: Creating Dance from User-Defined Labels.} Given the lack of user-tailored creative dance AIs, we propose ``PirouNet," a semi-supervised generative recurrent deep learning model that creates new dance sequences from choreographers' aesthetic inputs. We illustrate the use of PirouNet with categorical intensities of the Laban Time Effort, a staple of LMA's descriptive richness. Future users can choose different Laban-inspired labels, or define any subjective label they wish to use for creation, regardless of their knowledge of LMA. The artist then categorically labels a very small portion of an input dance dataset, typically of the order of 1\%. We provide a label augmentation tool that smartly increases this portion to about 25\%, in our case. Once trained, PirouNet creates new dances prompted by the artist's desired style. %As a by product, our semi-supervised method also enables automatic labeling of the dataset's unlabeled part. This proves interesting for future deep learning advances on dance that may benefit from fully labeled datasets which are still, at the moment, generally rare \cite{web_app_annotations}.

\begin{figure}
	\centering
 	\includegraphics[width=\linewidth]{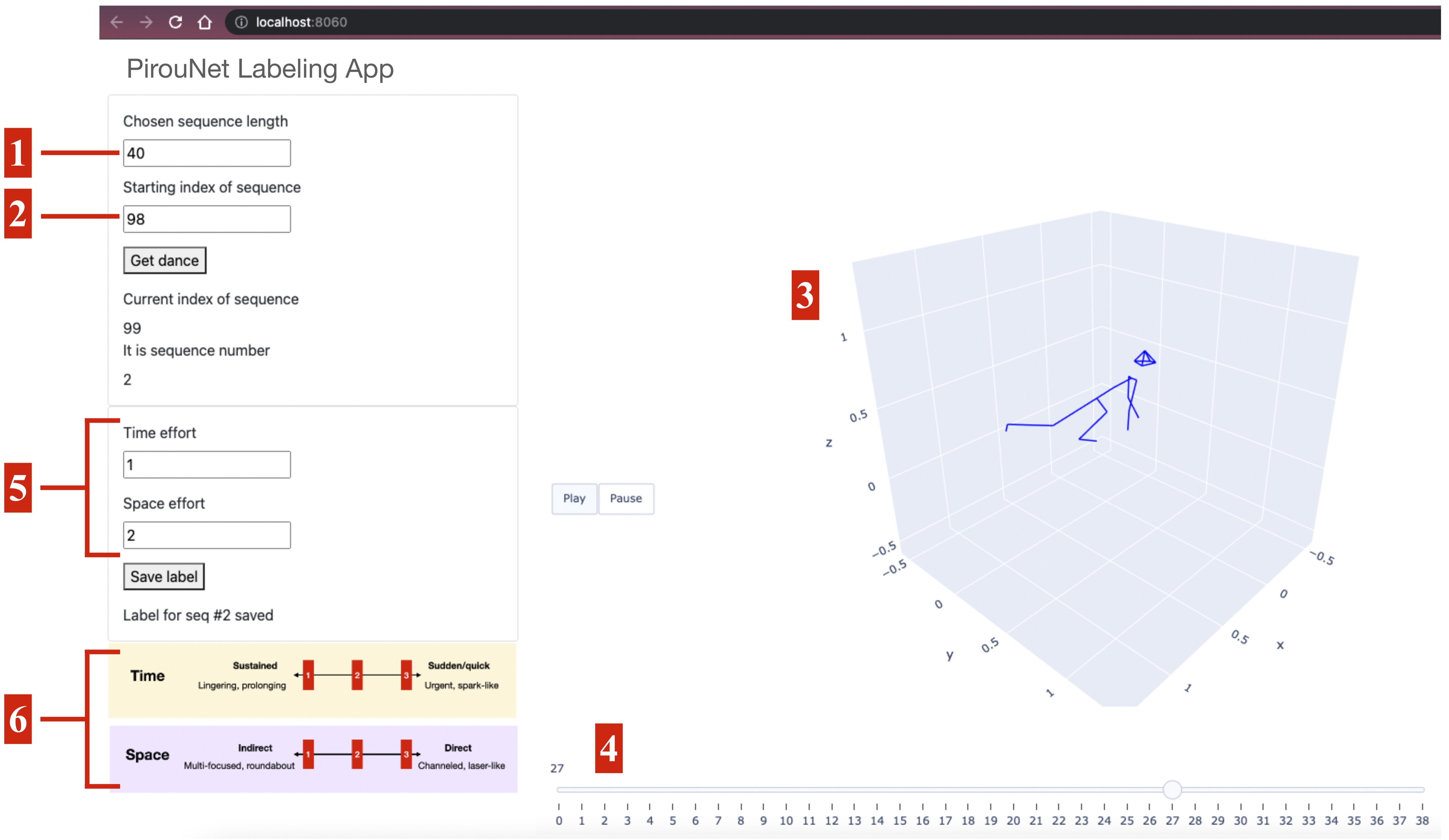}
	\caption{Screen capture of web labeling app. User enters amount of poses per dance sequence in (1), and the index of the first sequence's starting pose in (2). Upon clicking ``Get Dance," an animation of the fully connected skeleton appears in (3). The user can zoom and rotate the animation directly, and click into specific frames with (4). User enters Laban Effort (or any chosen label) in (5) and clicks ``Save label" to record the inputs to a CSV file. A spot is reserved in (6) for an infographic with instructions to ensure consistent labeling.}
	\label{fig:webapp}
\end{figure}

\paragraph{Contributions.} Our contributions are three-fold.
\begin{enumerate}
    \item We introduce the first deep generative model for dance based on LMA's Effort axis. %classes. % that is also adaptable to new choreographer-defined input classes. %Our model is comprised of a Long-Short Term Memory (LSTM) network and semi-supervised conditional variational autoencoder (SCVAE).
    \item We provide a novel dance labeling web application with a label augmentation toolkit, which distinguishes itself from currently available customizable annotation databases \cite{web_app_annotations} as it may be used on the user's own motion dataset. Fig. \ref{fig:webapp} shows an annotated screenshot of the app.
    \item We present a semi-supervised approach that, coupled with our web app, (i) allows the use of our model for dance generation based on \textit{any} class of the choreographer's choosing and (ii) limits manual annotation labor to 1\% of the dance dataset.
    %\item We leverage our method to also provide a large synthetic dataset of dance sequences labeled with Laban Efforts.
\end{enumerate}
%We hope that the advances showcased in this work will help democratize AI-powered dance generation and the use of LMA in deep learning models of motion.

\paragraph{Outline.} The paper is organized as follows. Section~\ref{sec:related} describes related work coupling LMA, deep learning and automatic generation of human motion and dance. Section~\ref{sec:methods} introduces the architecture and semi-supervised training of the proposed model, PirouNet. Section~\ref{sec:results} presents experimental results on dance creation from Laban Efforts, validated with a series of qualitative and quantitative metrics that demonstrate PirouNet's potential for AI-powered and choreographer-controlled dance generation. %Section~\ref{sec:discussion} discusses the impact of our work for the democratization of AI in dance and arts.

\section{Related Works}\label{sec:related}

We review related works on LMA and machine learning, as well as on the automatic creation of human motion and dance.

\paragraph{Predicting LMA from videos or pose sequences.}
A first class of learning methods focuses exclusively on predicting LMA from input dance videos. Researchers have successfully leveraged motion sensors  \cite{gesture_recognition} and pose detection systems \cite{kinect,automated} to automatically infer LMA labels from dance sequences, which subsequently allows assessing and improving movement techniques \cite{self_learning}. Recent advances in the field have extended the ability of these systems to generate the analyzed movement’s corresponding LN using feed-forward time-series neural networks \cite{ln_folk} and bi-directional RNNs \cite{zhang_automatic_2019}. %Similar work that applies  multi-class classifiers representing LN scores \cite{li_dance_2019}.

\paragraph{Predicting high-level dance information from LMA.}
Another class of methods uses existing LMA annotations (e.g. produced from the first class of methods) as inputs to perform downstream tasks, demonstrating the richness of LMA notations. For example, \cite{lma_folk_2015} can predict the similarity of two dancing motions from LMA. Supervised deep learning using LMA as inputs has also successfully extended computational movement analysis to emotion recognition \cite{emotion_exergames}, often relying on the complex temporal learning processes offered by Long Short Term Memory networks (LSTMs) \cite{Hochreiter97,wang_emotion_2020} and Convolutional Neural Networks (CNNs) \cite{emotional_analysis}.

\paragraph{Creating dance from LMA: non-deep AI and algorithmic approaches.}
%The aforementioned approaches either (i) predict LMA from input dances (for a downstream task), or (ii) extract high-level dance information from LMA input features. Noticeably, they do not use LMA to generate original dance choreography. 
LMA has played a key role in the emergence of a myriad of tools for generating dance directly from written scores \cite{wilke_2005} or vice versa \cite{gen_from_ln}. This approach's machine learning component mostly relies on associative memory \cite{laban_editor_AI} or look-up tables \cite{zhang_li_2006}, with an associated retrieval criterion. The subsequent adaptation of the dance motion is either implemented through hard-coded dynamic constraints or through a combination of such constraints and direct user-inputs. While such methods testify of the interest in generating dance from LMA, they are rather restrictive in the creative context of dance generation. Carlson et al. \cite{catalyst} make use of a genetic algorithm to generate static dance poses from Laban Efforts, meant to catalyze choreographic creation. 

\paragraph{Creating dance from non-LMA inputs: deep learning approaches.} Outside of the LMA community, researchers have introduced deep learning tools into the choreographic creative process. Most of these tools generate sequences of movement using variants of Recurrent Neural Networks (RNNs) \cite{og_rnns} to represent the time-component of the dance dynamical system, and (variational) autoencoders \cite{Kingma2014vae} to transform abstract latent variables into dance movements. A first class of algorithms generates choreography from scratch: the generated dance is completely new movement \cite{berman_from_scratch_2015,from_scratch_li}, without using any input from the choreographer. Other instances of innovative techniques achieving this include self-organizing maps \cite{dance_partner_SOM} and autoencoders for reacting to live movement \cite{james_2018_autoencoder_live}. Recent research has also introduced variational autoencoders \cite{pettee_2019} that encode sequences towards and generate sequences from a lower dimensional latent space. This enables generation of variations on given sequences as well as sampling new sequences. Lastly, other methods prompt sequences with music \cite{aist_music,lee_2019,alemi_2017,from_music_transformer}, which a choreographer can use to influence dance outputs. In all of these cases, the choreographer's creative control over the movement output is either limited to the choice of training data or that of an external prompt. To the best of our knowledge, there exists no model able to create dance sequences from choreographer-specific aesthetic labels or Laban Efforts. %In all of these cases, the choreographer's creative control over the movement output is limited, either depending on a choice of training set, centering around an abstract latent space from which it is difficult to defer meaning, or depending on a choice of prompt. 

% Dance generation
\paragraph{Creating movement from non-LMA inputs: conditional deep learning approaches.}

%Specifically, existing approaches produce choreography from inputs that are either (i) unconstrained or (ii) too constrained, without allowing the choreographers the option to tune which dance will be created, i.e. reducing the freedom of creation.

Recent methods, called label-conditioned movement generations, create human motion based on a discrete set of labels. To our knowledge, only two label-conditioned motion generation systems exist in the literature: Action2Motion \cite{action_2_motion} and Actor \cite{actor_vae}. Neither are adaptable to LMA-based motion generation nor to choreographer-guided dance generation, as they require a large, fully labeled dataset. Both Action2Motion and Actor train their methods from motion capture (MoCap) datasets, including the NTU dataset (containing over 100k movements) \cite{ntu_dataset} or the UESTC dataset (containing over 25k movements) \cite{uestc_dataset} labeled with human actions such as walking or throwing. Such labeled MoCap datasets are imbalanced in categories or noisy (inspiring \cite{action_2_motion} to produce their own action dataset) and do not consider dance moves nor LMA labels. Certain annotated dance databases exist \cite{aist_database,kpop_database,web_app_annotations}, although they are limited to their producers' specific genres, styles, and creative processes. We postulate that the lack of creative, yet precise, dance generation algorithms come from a lack of large dance datasets with labels meaningful to choreography as an artistic practice. Coming up with such a dataset for fully supervised models is especially difficult due to the time-consuming nature of such annotations.

\section{Methods}\label{sec:methods}

\begin{figure}
	\centering
 	\includegraphics[width=0.9\linewidth]{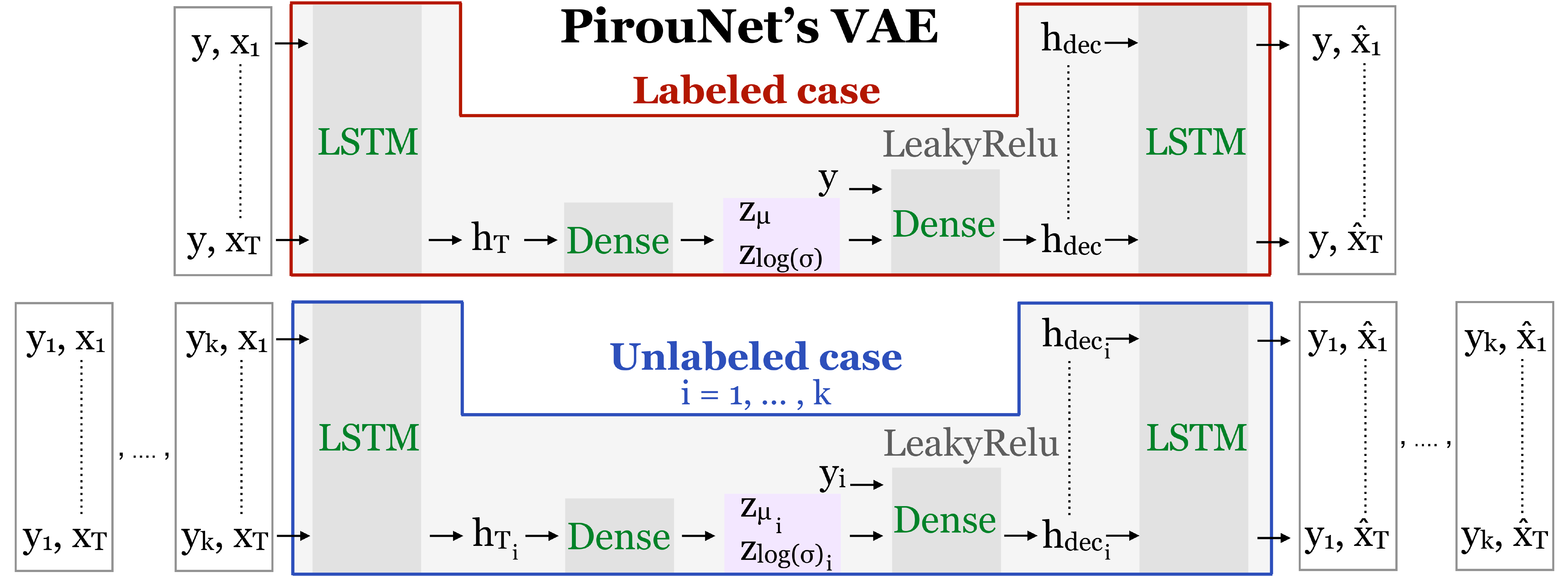}
	\caption{PirouNet's supervised and unsupervised training. A dance pose at time $t$ is represented by an input $x_t$, and a dance sequence is represented by an array of such poses, $x = [x_{\text{1}}, ..., x_T]$. The last hidden variable $h_T$ in the LSTM output is encoded into a continuous latent variable $z$. The decoder's LSTM takes in a hidden variable $h_\text{dec}$ repeated $T$ times. The Laban Effort $y$ associated to a pose $x_t$ is a categorical latent variable that represents the input label for that particular sequence. There are $k = 3$ distinct possible classes for the label $y$. Importantly, the VAE in the unlabeled case reconstructs every dance sequence $k$ times, using a different label value each time.}
	\label{fig:architecture}
\end{figure}

This section presents the proposed conditional dance generative model and the core components of PirouNet's architecture: a dance encoding and generative model that uses (i) a variational autoencoder (VAE) \cite{Kingma2014vae} inspired from \cite{pettee_2019}, coupled with (ii) motion dynamics through a LSTM \cite{Hochreiter97} network. In order to minimize the need for manual labeling, PirouNet also leverages (iii) a semi-supervised learning approach \cite{Kingma2014dgm}. PirouNet's source code implementing this model is available at \href{www.github.com/bioshape-lab/pirounet}{github.com/bioshape-lab/pirounet}. As PirouNet accepts both labeled and unlabeled input data for training, its VAE adopts two different forms, showcased in Fig. \ref{fig:architecture}. The VAE makes use of a linear classifier for the case of unlabeled data.
% The complete architecture is shown in Fig. \ref{fig:architecture_overview}.

% \begin{figure}
% 	\centering
%  	\includegraphics[width=\linewidth]{figs/arch_overview.pdf}
% 	\caption{PirouNet Architecture. A conditional encoder and generator are built off of a VAE. Wrapping LSTM layers ensure preservation of temporal dynamics.}
% 	\label{fig:architecture_overview}
% \end{figure}

\subsection{Conditional Dance Generative Model}

We propose the following conditional dance generative model:
\begin{equation}\label{eq:genmodel}
p_\pi(y)
=
\operatorname{Cat}(y \mid \pi); 
\quad 
p(z)
=
\mathcal{N}(z \mid 0, I) ; \quad 
p_{\theta}(x \mid y, z; \theta)
=
\mathcal{N}(f_\theta(y, z), \sigma^2),
\end{equation}
where $z$ is a continuous latent variable representing the dynamics - i.e. which movement is performed - and $y$ is a categorical variable denoting the choreographer specific or LMA-based label - i.e. how the movement is performed. If no label is available, $y$ is treated as a latent variable. The observed variable $x$ represents a dance sequence. The prior on $y$ is modeled as a multinomial distribution (chosen as a uniform categorical distribution in practice) $\operatorname{Cat}(y \mid \pi)$ with parameter $\pi$, while the prior on $z$ is represented by $\mathcal{N}(z \mid 0, I)$, the multivariate standard Gaussian following the VAEs' classical prior. The latent variables $y, z$ are assumed to be marginally independent. The forward generative model $p_{\theta}(x \mid y, z)$ is a Gaussian distribution parameterized by a mean $f_\theta(y, z)$ defined as a non-linear transformation of the latent variables $y, z$, and a variance $\sigma^2$. In practice, the non-linear transformation $f_\theta(y, z)$ is implemented as a neural network with weights $\theta$ called the decoder $f_\theta$.

The decoder further implements the dance dynamics through a series of temporal updates on the hidden variable $h^t$:
\begin{equation}
    h^{(t+1)} = \operatorname{Dyn}_\theta(h^{(t)}, z, y), \quad t=1, ..., T-1
\end{equation}
producing $T$ poses that are concatenated to produce a dance sequence. In practice, PirouNet uses LSTM layers to represent the dynamic operator $\operatorname{Dyn}_\theta$ \cite{Hochreiter97}.

%f(y, z) = [h1, ..., h_T-1] = [\operatorname{Dyn}_\theta(h_0, y, z), ..., \operatorname{Dyn}_\theta(h_t, y, z), ... \operatorname{Dyn}_\theta(h_{T-1}, y, z)]$

\subsection{Semi-Supervised Conditional Variational Autoencoder}

Based on the conditional dance generative model of Eq.~\eqref{eq:genmodel}, PirouNet solves the inverse problem of inferring the latent variables of movement and intention ($z, y$) from a dance sequence ($x$). PirouNet is built with a conditional variational autoencoder (VAE) architecture \cite{Sohn2015cvae,Kingma2014vae}, modified following the semi-supervised approach proposed in \cite{Kingma2014dgm} in order to include the (only partially observed) labels of movement intentions, as shown in Fig.~\ref{fig:architecture} and described below.

\paragraph{Encoder: Posterior of the movement latent variable $z$ (labeled and unlabeled cases.)} Inference in (conditional) VAEs involves computing an approximate posterior $q_\phi$ for the continuous latent variable $z$, which writes:
\begin{equation}\label{eq:qphi}
    \begin{cases}
      q_\phi(z|x, y) \quad \text{when a label $y$ is available,}\\
      q_\phi(z|x) = \sum_{y=1}^k q_\phi(z, y|x) = \sum_{y=1}^k q_\phi(z|y, x) \cdot q_\phi(y|x) \quad \text{otherwise}.
    \end{cases} 
\end{equation}
Both labeled and unlabeled cases in Eq.~\ref{eq:qphi} require the computation of the distribution $q_\phi(z| y, x)$, which is modeled as a multi-variate diagonal Gaussian distribution, as is standard in VAEs \cite{Kingma2014vae}. The role of the encoder $g_\phi$ is to output $q_\phi(z| y, x)$. In the unlabeled case, the encoder is used $k$ times (see Fig. \ref{fig:architecture}): once for each possible value of the label $y$ within the sum defining the unlabeled case in Eq.~\ref{eq:qphi}. Specifically, the encoder $g_\phi$ takes a sequence $x$ and a label $y$ as inputs and outputs the mean $\mu_\phi(x, y)$ and variance $\sigma^2_\phi(x, y)$ of the amortized approximate posterior $q_\phi(z|x, y)$, written as:
\begin{equation}
    q_\phi(z|x, y)
    = \mathcal{N}(g_\phi(x, y)) = \mathcal{N}(\mu_\phi(x, y), \sigma^2_\phi(x, y)).
\end{equation}
Importantly, the encoder extracts this posterior from the dynamics of the input sequence $x$, which is represented as:
\begin{equation}
    h^{(t+1)} = \operatorname{Dyn}_\phi(h^{(t)}, x, y), \quad t=1, ..., T-1,
\end{equation}
where the last hidden variable $h_T$ is fed to fully-connected layers to give $\mu_\phi(x, y)$ and $\sigma^2_\phi(x, y)$. In practice, just as for the decoder, this is implemented with LSTM layers \cite{Hochreiter97}.

\paragraph{Classifier: Posterior of the intention latent variable $y$ (unlabeled case).} The unlabeled case requires the additional computation of the posterior $q_\phi(y|x)$ of the categorical latent variable $y$ in Eq.\eqref{eq:qphi}. This is modeled as a multinomial distribution, as is standard in multi-class classification. Compared to a traditional VAE, we use an additional encoder-classifier (called classifier, for simplicity) whose role is to output $q_\phi(y|x)$. Specifically, the classifier takes a sequence $x$ as input and outputs the $k$ probabilities of the $k$ possible values for the label $y$ in the probability vector $\pi_{\phi}(x)$, i.e:
\begin{equation}
    q_\phi(y|x) = \operatorname{Cat}\left(y \mid \pi_{\phi}(x)\right).
\end{equation}
In practice, the classifier is implemented with fully-connected layers (Fig. \ref{fig:architecture}).

\paragraph{Reparametrization trick.} The VAE reparameterization trick is used to generate a sample $z$ from the approximate posterior $q_\phi(z|y, x)$, i.e. an element of the VAE's latent space that corresponds to a low-dimensional representation of the dance sequence $x$. We emphasize that $z$ represents a dynamic \textit{sequence}, and not a static dance pose --- which is crucial for the encoding of Laban Efforts that intrinsically characterize dynamics. Specifically, $z$ should be understood as a parameter of dynamics (e.g. describing the movement of a jump or spin), which is fed to the generative model's LSTM layers.

\paragraph{Decoder/Generator: Conditional Dance Generative Model.} The VAE's decoder corresponds to the implementation of the conditional dance generative model from Eq.~\eqref{eq:genmodel}. The decoder takes the encoded dance sequence $z$ (sampled with the reparametrization trick) and an assigned label $y$ as inputs. During unsupervised training, the decoder is used $k$ times (see Fig. \ref{fig:architecture}): once for each possible value of the label $y$. The decoder then outputs a reconstructed dance sequence $\hat x=f_\theta(z, y)$. To generate original dance choreography, a new random latent variable $z$ is sampled from an approximation of the marginal distribution $q_\phi(z|y)$ - corresponding to the body motion - where the label $y$ is chosen by the user.

% Once trained, the decoder plays the role of a ``generator" that can create new dances based on any movement variable $z$ and intention of movement (or choreographer-defined) label $y$.

%. The VAE's encoder performs inference, and is implemented as a neural network $g_\phi$ with weights $\phi$ also called the inference network or the recognition network. The unlabeled case requires the use of an additional network $h_\phi$ with weights $\phi$

\subsection{Semi-Supervised Training and Loss Function}

The training uses the semi-supervised framework and loss function prescribed in \cite{Kingma2014dgm} which considers the labeled and unlabeled cases separately. %Detailed derivations of the lower bounds are provided in the appendices. 

\paragraph{Labeled Case.} In the labeled case, our objective is to maximize the log-likelihood $\log p_\theta(x, y)$ to learn the parameters $\theta$ of the conditional dance model. This is traditionally performed via the maximization of its lower bound $-\mathcal{L}(x, y)$:
% \begin{align*}
% \log p_{\theta}(x, y) 
% & \geq \mathbb{E}_{q_{\phi}(\mathbf{z} \mid x, y)}\left[\log p_{\theta}(x \mid y, \mathbf{z})+\log p_{\theta}(y)+\log p(\mathbf{z})-\log q_{\phi}(\mathbf{z} \mid x, y)\right]\\
% & =-\mathcal{L}(x, y).
% \end{align*}
\begin{align*}
\log p_{\theta}(x, y) 
& \geq \mathbb{E}_{q_{\phi}(z \mid x, y)}[\log p_{\theta}(x \mid z, y)+\log p_{\pi} (y)]-\operatorname{KL}(p(z) \| q_{\phi}(z \mid x, y))
\\
& =-\mathcal{L}(x, y),
\end{align*}
where $\operatorname{KL}$ denotes the Kullback-Leibler (KL) divergence. In our case, the prior $p_{\pi} (y)$ is uniform over all labels, and the distribution $p_{\theta}(x \mid z, y)$ is modeled as a Gaussian. Thus, the quantity $\mathcal{L}(x, y)$ resembles a regularized $L_2$ reconstruction loss of a VAE with a modified KL divergence. The $L_2$ reconstruction loss is averaged over each body joint in a pose, and over each pose per sequence. 

\paragraph{Unlabeled Case.} In the unlabeled case, $y$ is missing and treated as another latent variable, in addition to $z$, over which we perform posterior inference. In this case, the objective is to maximize the marginalized log-likelihood $\log p_\theta (x)$, via the maximization of its lower bound $-\mathcal{U}(x)$ written as: 
\begin{align*}
\log p_{\theta}(x) 
& \geq 
    \mathbb{E}_{q_{\phi}(y, z \mid x)}\left[
        \log p_{\theta}(x \mid y, z)
        + \log p_{\theta}(y) 
        + \log p(z)
        - \log q_{\phi}(y, z \mid x)
    \right] \\
& = 
    \sum_{y=1}^k q_{\phi}(y \mid x)(-\mathcal{L}(x, y))+\mathcal{H}\left(
        q_{\phi}(y \mid x)
    \right) =     -\mathcal{U}(x),
\end{align*}
where $-\mathcal{H}$ denotes a loss term participating to the entropy:
\begin{equation}
    H(q_{\phi}(y \mid x)) = - \sum_{y=1}^k q_{\phi}(y \mid x) \log q_{\phi}(y \mid x) = -  \sum_{y=1}^k q_{\phi}(y \mid x) \mathcal{H}\left(
        q_{\phi}(y \mid x)
    \right).
\end{equation}

The loss $-\mathcal{H}\left(
        q_{\phi}(y \mid x)
    \right)$ and the regularized reconstruction loss, $\mathcal{L}(x, y)$, are now weighted by the confidence associated to each label $y$, $q_{\phi}(y \mid x)$.

\paragraph{Loss Function.} The training uses the loss function computed from a lower bound that encompasses the labeled and unlabeled cases:
\begin{equation}
\text{loss} = \sum_{(x, y) \sim \widetilde{p}_{l}}
    \mathcal{L}(x, y)
    + 
    \sum_{x \sim \widetilde{p}_{u}}
    \mathcal{U}(x) + \alpha \cdot \mathbb{E}_{\widetilde{p}_{l}(x, y)}\left[-\log q_{\phi}(y \mid x)\right],
\end{equation}
where $\tilde p_l(x, y)$ and $\tilde p_u(x)$ are the empirical distributions over the labeled and unlabeled subsets, and the third term is a classification loss weighted by the hyperparameter $\alpha$ which controls the relative weight between dance generation and classification objectives \cite{Kingma2014dgm}.

\section{Results}\label{sec:results}

We qualitatively and quantitatively evaluate PirouNet for semi-supervised classification and generation of choreography. During experiments, we find that these two goals are best achieved with different hyperparameters and levels of supervision. Therefore, we present a model primarily trained for the purpose of creating novel and conditional dance (PirouNet\textsubscript{dance}), as well as a model that achieves better classification accuracy (PirouNet\textsubscript{watch}).

Like most deep learning tools for dance, we leverage motion data in keypoint format. This format, often included in large movement datasets \cite{aist_database,ntu_dataset,hdm05_database}, represents the body as a cloud of 3D points, each representing a unique joint. Available and high-performing pose-estimation software \cite{deeppose,stacked_pose,multi_pose} makes keypoint format accessible to smaller, homemade datasets as well. We use half of Pettee's keypoint dataset \cite{pettee_2019}, featuring a trained contemporary dancer in solo improvisation. We choose this particular style of dance for its predisposition to LMA classification, a tool developed for Western styles of dance. Future work is needed to validate PirouNet for other forms of dance and annotations. In this spirit, we propose a Dash \cite{dash_plotly} app, pictured in Fig. \ref{fig:webapp}, for easy manual labeling of any keypoint dataset.

\subsection{Datasets}

The dataset \cite{pettee_2019} is comprised of 36,396 poses extracted from six uninterrupted dances captured at a rate of 35 frames per second. This amounts to about 20 minutes of real-time movement of an experienced dancer. Each pose features 53 joints captured in three dimensions, normalized such that the dance fits within inside a unit box. The dancer's barycenter is fixed to one point on the (x,y) plane. From the pose data, we extract 36,356 sliding sequences of 40 continuous poses, and manually label 350 of these sequences (0.96\% of the dataset) which do not share any of the same poses. This takes an experienced dancer (the principal author) about 3 hours, identifying if the movement's Laban Time Effort is Low, Medium, or High. We apply two automated techniques to augment this labeling to 9,167 labeled sequences (representing 25.2\% of our unlabeled dataset) in total, split between 45\% Low, 34\% Medium, and 21\% High Efforts. (i) We automatically label all sequences between sequences that share a same Effort. For example, if two back-to-back sequences are deemed to have Low Time Effort, all sequences that are a combination of the poses in these two sequences are also labeled with Low Time effort. (ii) We extend every label to all sequences starting within 6 frames (0.17 seconds) before or after its respective sequence.

\subsection{Training}

All experiments are built using the PyTorch library \cite{pytorch} and run on a server with two Nvidia A30 GPUs and two CPUs, each with 16 cores. We train using an Adam optimizer with standard hyperparameters \cite{adam_opt}. We present results for the PirouNet architecture resulting from a hyperparameter search using Wandb \cite{wandb} on batch size, learning rate, number of LSTM and dense layers, as well as hidden variable sizes. PirouNet uses 5 LSTM layers with 100 nodes in both the encoder and the decoder. The classifier features 2 ReLU-activated \cite{relu} linear layers with 100 nodes. The latent space is 256-dimensional, which is approximately 25 times smaller than the 6360-dimensional initial space. We train for 500 epochs with a learning rate of $3\text{e}^{-4}$ and a batch size of 80 sequences. We select different epochs for PirouNet\textsubscript{dance} and PirouNet\textsubscript{watch} to minimize validation loss. For unsupervised training, we use 35,538 40-pose sequences,
with the remaining sequences being reserved for testing. For supervised training, PirouNet\textsubscript{dance} and PirouNet\textsubscript{watch} are trained on 79\% (16.6\% of entire training set) and 92\% (18.8\% of entire training set) of the labeled sequences, respectively. We reserve 5\% of the labeled sequences for validation, and 3\% for testing. 

\subsection{Semi-supervised Classification of Laban Efforts}

\begin{figure}
	\centering
 	\includegraphics[width=0.9\linewidth]{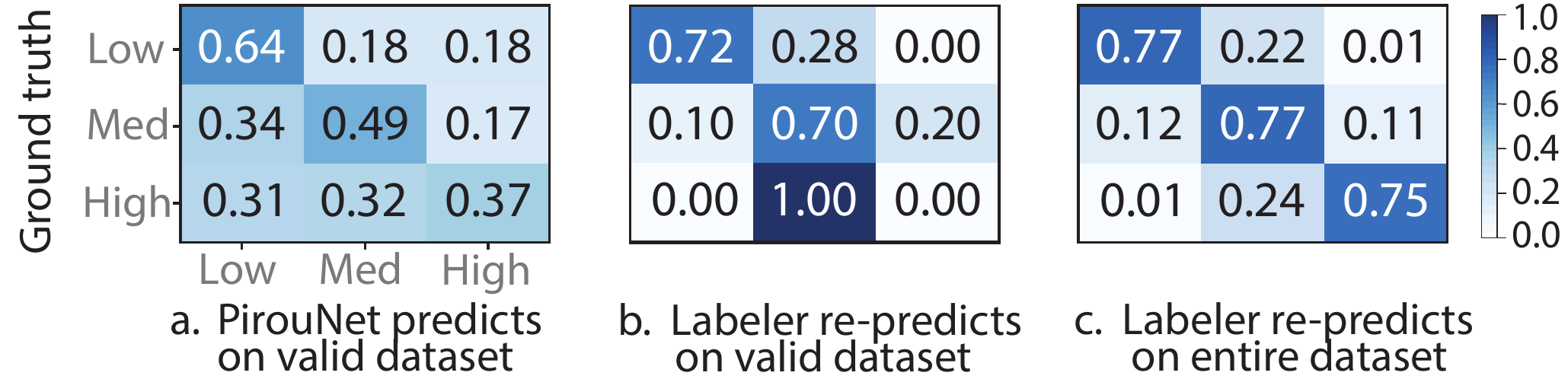}
	\caption{Confusion matrices for classification of validation dance sequences' Time effort as being ``Low", ``Medium", or ``High". a. PirouNet\textsubscript{watch} classifies the validation dataset. b. The labeler reclassifies a comparable validation dataset. c. The labeler reclassifies a comparable entire dataset (train, validation, and test).}
	\label{fig:confusion}
\end{figure}

While the primary purpose of this work is to generate Effort-specific dance, we examine classification performance to investigate training: we evaluate the accuracy of the classifier in attributing the correct categorical Time Effort to a dance. Because evaluating Laban efforts is subjective, the ground-truth labels provided by the human labeler may be considered ``noisy". To account for this in our evaluation of PirouNet\textsubscript{watch}, the human labeler also evaluates their own self-accuracy by relabeling the validation and entire datasets (Fig.~\ref{fig:confusion}(b, c)).

When classifying the validation set (Fig. \ref{fig:confusion}a), PirouNet\textsubscript{watch} succeeds with 50.1\% accuracy, which represents 75\% of the labeler's self-accuracy. In the case of test data, PirouNet\textsubscript{watch} is 72\% as accurate as the labeler. This is a satisfactory performance, considering the subjective nature of the task, the limited access to labeled data, and the observation that classifying the validation set was challenging even for the human labeler.

%The contrast with the average classification accuracy on training data, 85.3\%, indicates potential overfitting. 
%PirouNet\textsubscript{dance}, which is trained on fewer labels for more epochs, also exemplifies this behavior. %While it generates correctly labeled movement, it poorly classifies sequences, attributing 89\% of Low Effort and 83\% of High Effort sequences to the Medium Effort category. 

% \begin{figure}
% 	\centering
%  	\includegraphics[width=0.8\columnwidth]{figs/conf_test.pdf}
% 	\caption{Confusion matrices for classification of test dance sequences' Time effort as being "Low", "Medium", or "High". a. PirouNet classifies the test dataset (249 sequences). b. The labeler reclassifies a comparable test dataset.}
% 	\label{fig:test_confusion}
% \end{figure}

We hypothesize that results could be further improved by increasing the training dataset size, as well as optimizing the trade-off between training the VAE and the classifier. This would include examining the impact of the loss's control term $\alpha$ during later training. A hyperparameter search limited to early training shows $\alpha=0.1 \cdot \frac{n_{\text {unlabeled }}}{n_{\text {labeled }}}$ is optimal, where $n_\text{unlabeled}$ and $n_\text{labeled}$ are the unlabeled and labeled dataset sizes. Another potential solution would be switching to a regression setting. Using continuous scalar variables as labels would better reflect the continuous nature of the Effort spectrum and be better suited to deal with noise between Low and High Efforts. Fig \ref{fig:confusion}b shows an example of how the discontinuous categories can result in strong self-disagreement for humans, which affects the performance of the classifier on those same sequences (Fig \ref{fig:confusion}a). While selected for classification, PirouNet\textsubscript{watch}'s generation performance is also satisfactory.% (see appendices).

It is worth noting that the ground truth and relabeled sets of data are not exactly identical in labeling procedure: while the dance sequences are the same, a sequence that was manually classified in one dataset may have been classified via label augmentation in the other. We do not expect this to significantly impact training, as the augmentation procedure was designed to agree with manual labeling. However, this signifies that a misclassification of a small number of sequences by the labeler can propagate through augmentation, and result in 100\% misclassification for a given label, as is the case in Fig. \ref{fig:confusion}b.  We note that unlike PirouNet\textsubscript{watch}, the labeler almost never re-predicts a sequence to be more than one categorical bin away from the ground truth. This highlights the use of regression classification as a means of preserving information about the labels' meanings.  We leave the improvements discussed above for future iterations.

\subsection{Reconstructions of Choreographies}

\begin{figure}
	\centering
 	\includegraphics[width=\linewidth]{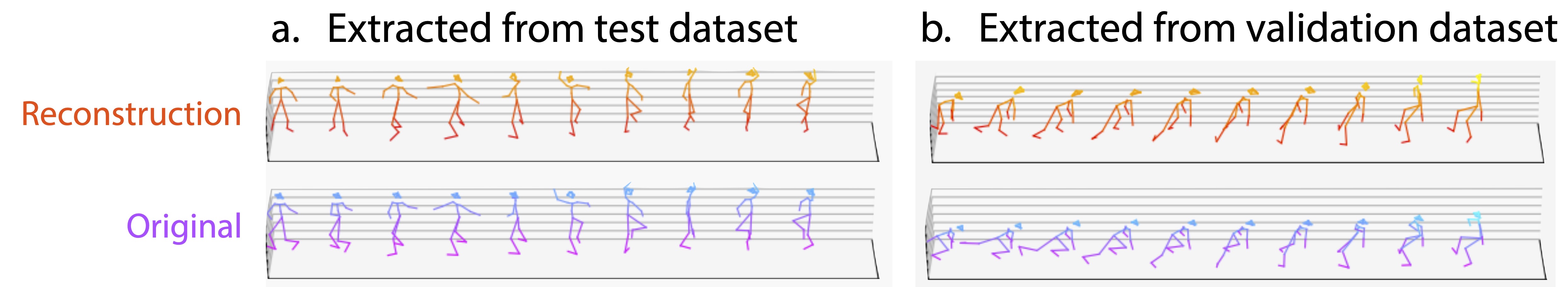}
	\caption{Reconstructions of dance sequences. a. Sequence extracted from test data featuring a fast pirouette. b. Sequence extracted from validation data featuring a multi-axial rotation on the floor with gradual leg and arm extension.}
	\label{fig:reconstructions}
\end{figure}

\paragraph{Qualitative assessment.} 
Artifacts obtained on validation and test datasets, two of which are depicted in Fig. \ref{fig:reconstructions}, show that PirouNet\textsubscript{dance} successfully encodes and decodes a given sequence with very little variation. Reconstructions capture complex movements that include rotations (``pirouettes"), changes in height, and changes in velocity. The Laban Time Effort is preserved, as indicated by the identical time evolution of the reconstructions. %We provide GIFs in the appendices. % (see appendices for GIFs).% GIFS of dance sequences are provided in the appendices.

% TODO: for a Run with 6% labeled data, show, qualitatively (i.e. visually) that the reconstructions work, i.e. on the validation set.
% - show input sequences of poses and output sequence of poses side to side
% - e.g. a few different movements (harold is on the floor, harold rotates, etc)
\paragraph{Quantitative assessment.}

To assess PirouNet\textsubscript{dance}'s ability to reconstruct movement, we use the root mean squared error representing the average joint distance (AJD) between original and reconstructed sequences, pose per pose. For N sequences with $T = 40$ poses of $J = 53$ joints each, the AJD writes as:
\begin{equation}
\bar{D}=\dfrac{1}{N \cdot T \cdot J} \sum_{n=1}^{N}\sum_{t=1}^{T}\sum_{j=1}^{J}\sqrt{\left|x_{j}^{t,n}-\hat{x}_{j}^{t,n}\right|^{2}+\left|y_{j}^{t,n}-\hat{y}_{j}^{t,n}\right|^{2} + \left|z_{j}^{t,n}-\hat{z}_{j}^{t,n}\right|^{2}}.
\end{equation}

For the validation dataset ($N = 455$), the average joint difference $\Bar{D}_{\text{validation}}$ is $2.8\text{e}^{-2}$. On the test dataset ($N=249$), the average difference is $\Bar{D}_{\text{test}} = 1.8\text{e}^{-2}$, which, remarkably, matches $\Bar{D}_{\text{train}}$ ($N=7,161$). Since the joint coordinates are scaled to fit inside a unit box, $\Bar{D}_{\text{validation}}$ and $\Bar{D}_{\text{test}}$ represent less than 2.8\% and 1.8\% of the dancer's height, 4.8 cm and 3.1 cm. 
While trained for classification purposes, PirouNet\textsubscript{watch} has a satisfactory reconstruction error, with $\Bar{D}_{\text{validation}} = 3.6\text{e}^{-2}$ and $\Bar{D}_{\text{test}} = 2.0\text{e}^{-2}$. %Fig. compares the pose-by-pose $d_\text{pose}$ for the reconstructed and original sequences featured in Fig. \ref{fig:reconstructions}.

% TODO: for a Run with 6% labeled data, show, quantitatively that the reconstruction work
% - give the reconstruction loss on the validation set 
% i.e. recon_loss here: https://github.com/bioshape-lab/move/blob/b3dc8904322594768aaf16ae125ee090a767f1ab/move/nn.py#L340
% and try to have it in units of cm (look at mean versus sum, etc)
% donner une reconstruction error en cm:
% i.e., en moyenne, a quelle distance sont les positions des joints reconstruits, par rapport aux joints inputs.

\subsection{Creation of Choreographies based on Laban Efforts}
% We find PirouNet's latent space to be entangled, meaning that the Laban Effort is encoded in the latent variable as a whole. This makes sense, as Laban Efforts and dynamics are not independent in the dataset: a movement with a low Effort has a higher probability of being a slow movement. In light of this, 
We propose conditionally generating dance by sampling from an approximation of the marginal conditional distribution $q_\phi(z|y)$ of the latent variable $z$, given an input Laban effort $y$. Effectively, this means sampling dance from neighborhoods in the latent space featuring a high density of previously encoded same-Effort sequences (manually or automatically labeled). Future work will focus on disentangling the latent space in order to enable conditional generation independently from previously encoded data. Outside of the conditional framework, PirouNet can generate dance in the general style of the training data by sampling random latent variables.

\paragraph{Qualitative assessment.} 
For each of the three labels (Low/Medium/High Effort), we conditionally generate 75 sequences from PirouNet\textsubscript{dance}'s marginal distribution. The assessment of this 225-sequence benchmark set is two-fold.

\begin{figure}
	\centering
 	\includegraphics[width=0.8\linewidth]{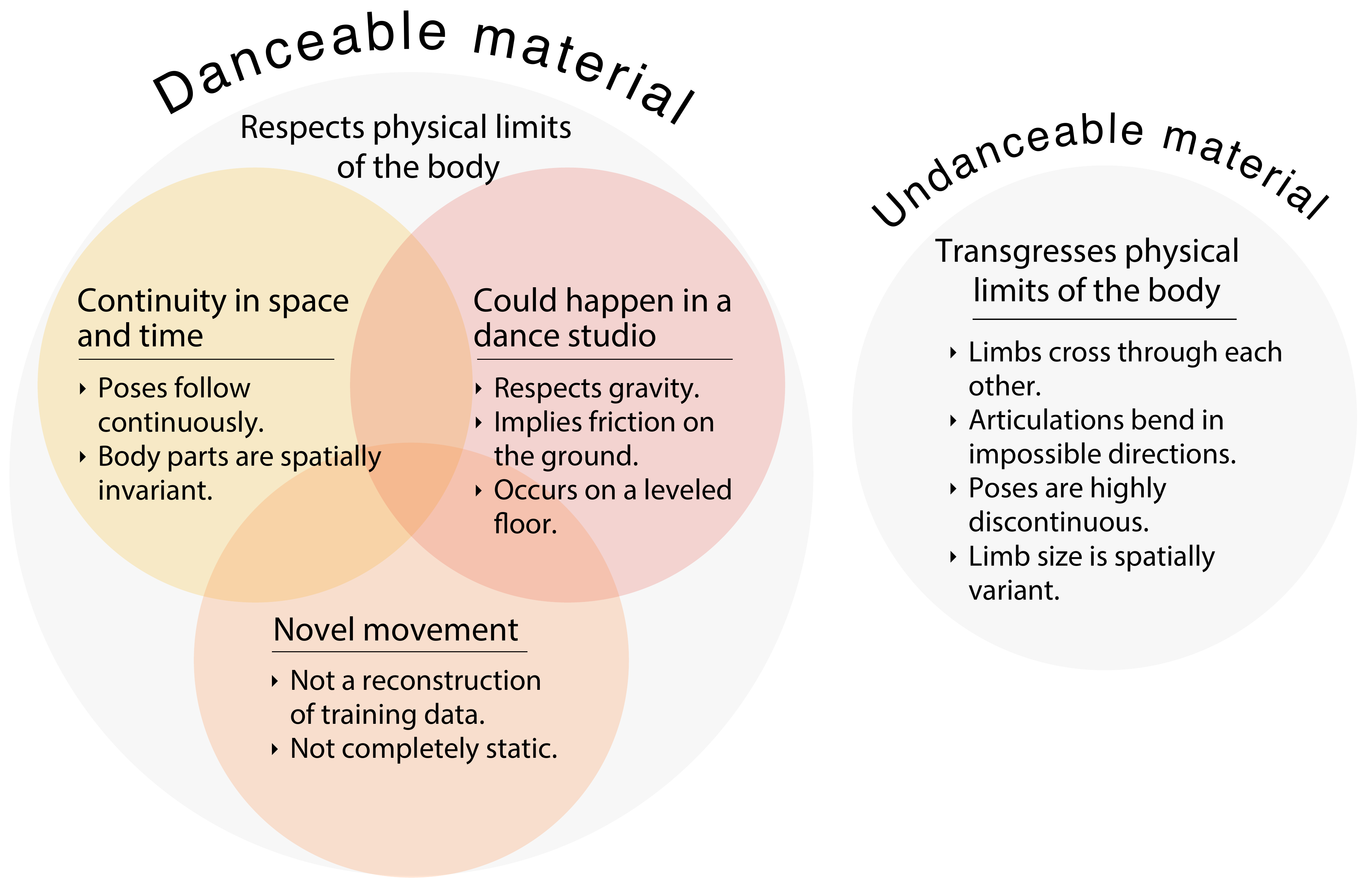}
	\caption{Infographic of danceable versus non-danceable material, as generated by PirouNet\textsubscript{dance}.}
	\label{fig:danceability}
\end{figure}

First, we examine the overall ``danceability" of the generated material. We determine danceable material as movement that (i) would be physically realizable by the training set's dancer in at least one physical environment and (ii) is largely continuous in space and time. Within this broad category, we find three sub-categories, outlined in Fig. \ref{fig:danceability}. Most of the danceable sequences belong to at least two of these sub-categories, which shows that PirouNet\textsubscript{dance} has learned to produce danceable materials.

\begin{figure}
	\centering
 	\includegraphics[width=\linewidth]{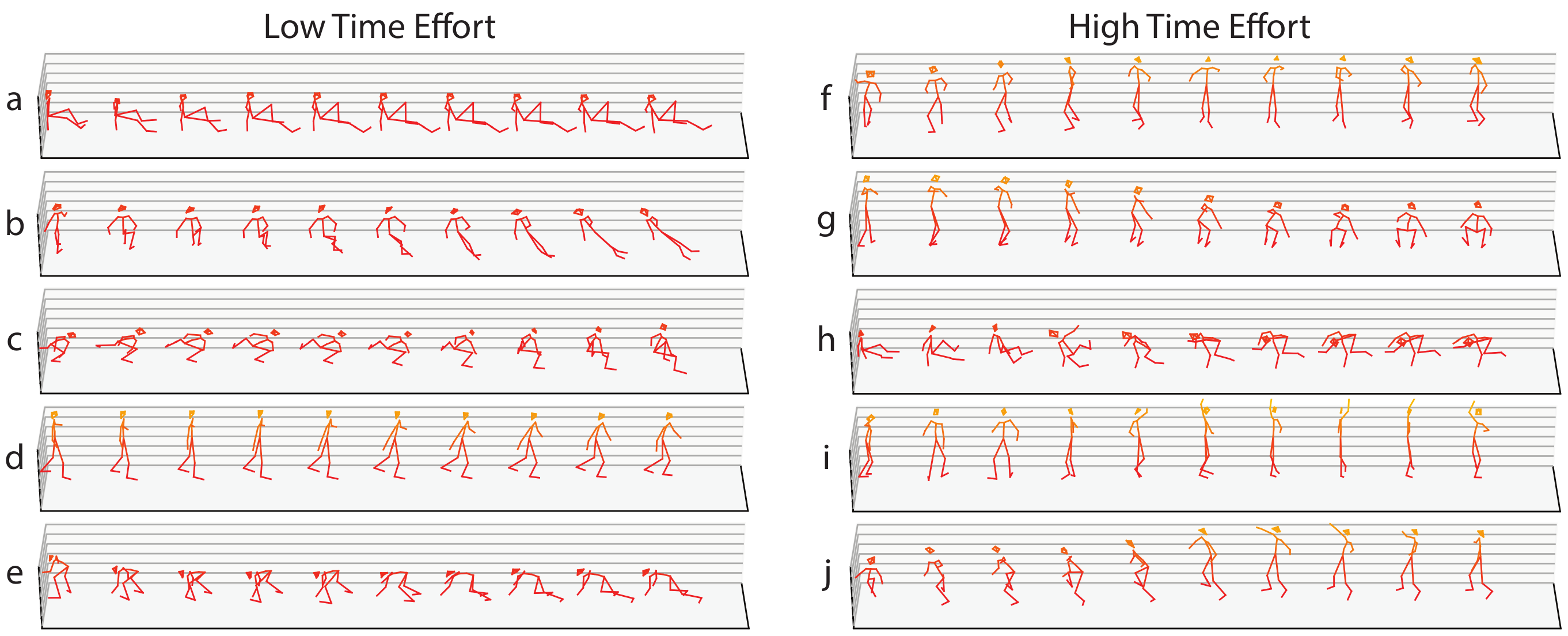}
	\caption{Low/High Time Effort dance sequences created by PirouNet\textsubscript{dance}. The labeler describes the experience of Time Effort as relating to the presence (or absence) of impulse, sense of urgency, and explosive momentum. a. Slow and continuous leg extension from V-sit. b. Continuous transition from crouched to plank position. c. Smooth weight transfer through deep plié. d. Forward standing weight transfer through small plié. e. Transition from standing to plank position through double rond-de-jambe. f. Demi-pirouette lead by elbow, ending in quick change in direction. g. Spontaneous drop to a crouched position. h. Leg swing leads a sharp torso rotation. i. Pirouette with arm thrown up. j. Explosive extension of arms and torso from plié.}
	\label{fig:high_low_effort}
\end{figure}

Second, we examine PirouNet\textsubscript{dance}'s display of Time Effort in the danceable material. Fig. \ref{fig:high_low_effort} depicts PirouNet\textsubscript{dance} originals grouped by intended Effort. Looking at the video format of such sequences, we remark that instances of discontinuity between poses, appearing as quick jumps, arise in a sequence's first or last five frames. The (danceable) movement is otherwise realistic, and often totally reproducible by an experienced dancer. We provide video examples at \href{www.github.com/bioshape-lab/pirounet}{github.com/bioshape-lab/pirounet}.

%The labeler describes the internalized experience of this effort as relating to impulse. A movement with High Time Effort is characterized by a sense of urgency that often materializes in explosions of momentum, small or large, tugging at the motion in unforeseeable ways. Fig. \ref{fig:high_low_effort} depicts PirouNet\textsubscript{dance} High Time Effort originals that agree with this definition. For its part, PirouNet\textsubscript{watch} sometimes fails to produce the sudden and quick quality of movement expected for High Time effort, instead producing slow writhing and wriggling on the ground, closely following some of the training sequences. 
% Moving towards the other end of the spectrum, the labeler describes Low Time Effort as intentionality for self-sustaining and self-governed movement. Low Time Effort may motivate controlled and predictable movement just as it may motivate comfortable, easy, and thoughtful explorations. In both cases, the motion is natural, flowing through a grounded but malleable body. Fig. \ref{fig:high_low_effort} demonstrates such sequences conditionally produced by PirouNet\textsubscript{dance}. In some cases, sequences completely miss the mark, adopting urgent and unpredictable gestures.%Similarly to some training sequences, some generated sequences feature juxtaposed Efforts, or no distinct Effort at all. We remark that instances of lack of continuity between poses (implying physically impossible transitions from pose to pose) are most common towards the first four or last four frames of a sequence.% TODO: for the same run with 6% labeled, show, qualitatively (visually) some sequences *created* by harold given a Laban effort, and explain why, intuitively they make sense.

\begin{figure}
	\centering
 	\includegraphics[width=0.6\linewidth]{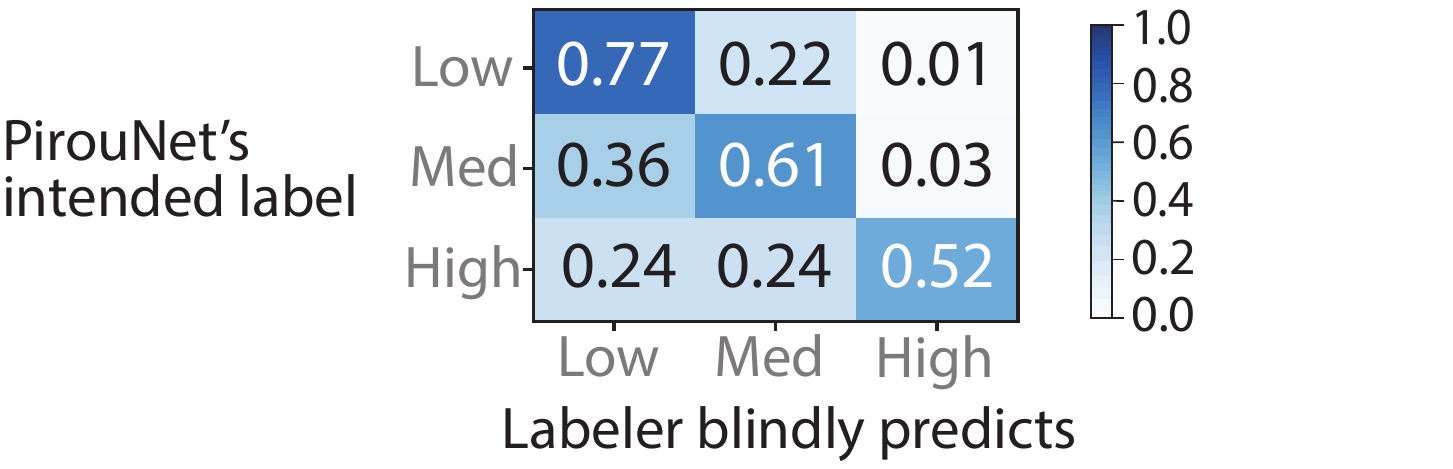}
	\caption{Confusion matrix on the danceable portion of the benchmark set (225 sequences). Generation is the least accurate for High Effort, possibly due to the smaller portion of High Effort labels in the training data. Notably, the human classifier correctly recognizes the majority of the PirouNet's intended Efforts.}
	\label{fig:blind_gen}
\end{figure}

\paragraph{Quantitative assessment.}
We evaluate PirouNet\textsubscript{dance}'s Effort-centric generative model using the 225-sequence benchmark set. The labeler identifies 96.0 \% of the dataset as danceable, which is 11.5\% better than a benchmark set randomly sampled from the latent space. Of the danceable portion, 98.1 \% of the conditionally sampled set respects the constraints of the physical environment, compared to 72.2\% of the randomly sampled set. Only 0.46\% of the former shows some non-physical discontinuity, such as a noticeable change in limb size, versus 8.8\% for the latter. Therefore, the marginal conditional method proves to be advantageous for the purposes of creating realizable and meaningful dance for the user, outside of providing Effort-specific movement. To determine PirouNet\textsubscript{dance}'s accuracy in generating appropriate Efforts, the labeler blindly labels the danceable portion of the benchmark set without knowledge of PirouNet\textsubscript{dance}'s intended Effort. On average, the human labeler correctly classifies 63\% of PirouNet\textsubscript{dance}'s sequences for a given Effort (Fig. \ref{fig:blind_gen}). This represents 83\% of the labeler's own self-agreement on the validation and training data (Fig. \ref{fig:confusion}c). %Additional results are reported in appendices.

\section{Conclusions}

This paper presents PirouNet, the first Laban Effort based deep generative tool for dance, made possible via a semi-supervised conditional dynamic variational autoencoder. While illustrated on Laban Efforts, our tool can be readily used with any choreographer-specific categorical labels. Our keypoint labeling web-app enables anyone to classify any keypoint data-set with any annotations, empowering artists to shape their own AI-tool. We demonstrate PirouNet's ability to classify the Laban Time Effort and generate new choreography based on this Effort as comparable to that of an experienced dancer. Conditional generation from the marginal distribution $q_\phi(z|y)$ offers largely accurate, diverse, and realizable choreography. Randomly sampling from the latent space at large also produces realistic and novel dance, providing an exciting choreographic tool outside of categorical generation. Future work will focus on improving classification accuracy and disentanglement of the latent space, as well as examining PirouNet's implementation and functionality for actual choreographic practices. We hope this artist-centric, adaptable tool will act as a launching pad for engaging with old repertoire and inspiring new choreography from a completely new creative standpoint. While demonstrated on dance, the proposed method can be extended to other forms of art creation, inspiring AI-based tools tailored to the style and intuition of their artist.

\bibliographystyle{splncs04}
\bibliography{move_bib}




% LNCS Homepage, \url{http://www.springer.com/lncs}. Last accessed 4
% \end{thebibliography}
\end{document}